\documentclass{article}

\usepackage[preprint]{neurips_2026}
\usepackage[utf8]{inputenc} 
\usepackage[T1]{fontenc}    
\usepackage{hyperref}       
\usepackage{url}            
\usepackage{booktabs}       
\usepackage{amsfonts} 
\usepackage{amsmath}
\usepackage{nicefrac}       
\usepackage{microtype}      
\usepackage{xcolor}         
\usepackage{graphicx}
\usepackage{enumitem}
\usepackage{array}
\usepackage{tabularx}
\usepackage{wrapfig}
\usepackage{makecell}
\usepackage{caption}
\usepackage[most]{tcolorbox}
\tcbuselibrary{listings,breakable}
\newtcblisting{promptlisting}[1]{
  enhanced,
  breakable,
  colback=gray!5,
  colframe=gray!55,
  title=#1,
  fonttitle=\bfseries,
  coltitle=black,
  boxrule=0.6pt,
  arc=2pt,
  left=6pt,
  right=6pt,
  top=6pt,
  bottom=6pt,
  listing only,
  listing options={
    basicstyle=\ttfamily\footnotesize,
    breaklines=true,
    columns=fullflexible,
    keepspaces=true,
    showstringspaces=false
  },
  before skip=8pt,
  after skip=8pt
}
\usepackage{algorithm}
\usepackage{algpseudocode}

\title{OPTScientist: Multi-Agent Discovery of Typed Optimizer Programs for Transformer Pretraining}

\author{%
  Zhongzheng Li$^{1,2,3}$,
  Tiancan Feng$^{3}$,
  Wenhao Li$^{3}$,
  Qingsong Ran$^{3}$,\\
  \textbf{
  Shikun Feng$^{3}$,
  Xiaoyuan Zhang$^{3}$,
  Yue Wang$^{3}$,
  Xiaoguang Zhao$^{1}$ }\\
  $^{1}$Institute of Automation, Chinese Academy of Sciences \\
  $^{2}$University of Chinese Academy of Sciences \\
  $^{3}$Zhongguancun Academy \\
    \texttt{zhangxiaoyuan@bza.edu.cn}
}

\begin{document}

\maketitle

\begin{abstract}
Designing optimizers for modern deep learning remains a challenging scientific problem, requiring the joint consideration of optimization geometry, state dynamics, numerical stability, implementation constraints, and empirical generalization. Existing automated optimizer discovery methods typically search either over unconstrained code spaces or within narrowly parameterized optimizer families. The former is flexible but often produces invalid or uninterpretable programs, while the latter is stable but limits novelty. We introduce OPTScientist, a theory-guided multi-agent framework for optimizer discovery in a typed domain-specific language (DSL). OPTScientist formulates optimizer design as a constrained scientific search process, where candidate updates are expressed through direction, scaling, preconditioning, regularization, state, and grouping modules. Four role agents, Theorist, Designer, Engineer, and Reviewer, collaborate within a single orchestration loop to propose hypotheses, synthesize DSL candidates, compile and evaluate optimizers, and critique results. To overcome the limitations of a fixed search space, OPTScientist combines evolutionary search over optimizer programs with a second-stage mechanism that proposes small DSL extensions when repeated failures reveal representational bottlenecks. Using this framework, we discover RS-MR, a reduced-state matrix optimizer that improves transformer pretraining over strong baselines under our native evaluation protocol. Our results suggest a path toward automated optimizer science grounded in theory, typed programs, compiler validation, and closed-loop experimentation.

\end{abstract}

\section{Introduction}

Optimizers are central to modern deep learning systems \citep{kingma2014adam, loshchilov2019decoupled}. They determine training speed, stability, memory consumption, final validation performance, and the feasibility of scaling models under fixed compute budgets. Despite substantial progress, optimizer design remains largely expert-driven \citep{zhao2024galore, liu2023sophia, vyas2024soap, zhang2024adam, liu2025muon}. Methods such as AdamW, Lion, Muon, and Shampoo encode different assumptions about update geometry, state accumulation, scaling, preconditioning, and regularization, but these assumptions are usually introduced through a mixture of intuition, partial theory, and extensive empirical iteration.

The difficulty is that an optimizer is not merely a formula mapping gradients to updates. It is a coupled dynamical system involving stochastic gradients, historical states, adaptive scales, numerical safeguards, parameter grouping rules, and implementation constraints. Small design changes can substantially alter convergence and stability. As a result, the optimizer design space is large, but only a small subset is scientifically meaningful, computationally feasible, and robust under realistic training protocols \citep{liu2025cosmos,wen2025fantastic}.

Automated optimizer discovery offers a promising alternative, but existing approaches face a tension between expressivity and reliability \citep{boiko2023autonomous,baek2025researchagent}. Unconstrained program search or language-model-based code generation can explore broad update rules, but often produces invalid, unstable, redundant, or difficult-to-interpret candidates. Conversely, searching within a fixed optimizer family improves reliability, but restricts discovery to local variants of known methods. What is missing is a framework that preserves expert-like theoretical priors while making optimizer design executable, searchable, auditable, and extensible \citep{ghafarollahi2025sciagents,hong2023metagpt}.

We introduce \textsc{OPTScientist}, a theory-guided multi-agent framework for discovering optimizers as typed programs. Rather than generating arbitrary Python code, \textsc{OPTScientist} searches within a typed optimizer domain-specific language (DSL), where candidate updates are expressed through direction, scaling, preconditioning, regularization, state, and grouping modules. A compiler parses each candidate, checks tensor and scalar compatibility, lowers it into an intermediate representation, and generates an executable optimizer. This compiler-backed representation rejects many invalid candidates before training and keeps discovered optimizers compact and inspectable.

At the search level, \textsc{OPTScientist} organizes optimizer discovery as a scientific loop. A \textsc{Theorist} proposes search hypotheses and priors from shared memory, a \textsc{Designer} converts them into diverse DSL candidates, an \textsc{Engineer} compiles, repairs, and evaluates candidates, and a \textsc{Reviewer} scores them for novelty, validity, performance, target-stage behavior, and scalability. These roles are coordinated within a single orchestration loop rather than as independent processes. The resulting workflow turns optimizer discovery from blind mutation into an iterative process of hypothesis generation, program synthesis, execution, critique, and revision.

A key component of \textsc{OPTScientist} is two-stage evolution. In the first stage, an evolutionary algorithm searches for optimizer programs within the current DSL. In the second stage, when repeated failures, target-stage plateaus, or compiler bottlenecks reveal representational limitations, the system proposes conservative DSL extensions, such as safe macro-like composition primitives. Thus, \textsc{OPTScientist} evolves not only optimizer candidates but also, when necessary, the language used to express them.

Using this framework, we discover \textbf{Reduced-State MAGMA-RowNorm} (\textbf{RS-MR}), a low-state matrix optimizer for transformer weight matrices. RS-MR combines row-normalized polar directions, lightweight global damping, and soft blockwise update gating. Unlike heavier preconditioned methods, it improves matrix update selectivity with only a compact block-level state. In our native transformer pretraining benchmark, RS-MR consistently outperforms strong baselines such as Muon while remaining close to Muon's optimizer-state memory footprint.

Our contributions are as follows:
\begin{itemize}
    \item We propose a typed-program formulation of automated optimizer discovery, where candidate optimizers are expressed in a compiler-backed DSL rather than generated as unrestricted Python code.

    \item We develop \textsc{OPTScientist}, a theory-guided multi-agent framework that coordinates hypothesis generation, DSL candidate synthesis, compilation, evaluation, reviewer scoring, and conservative DSL evolution within an auditable scientific search loop.

    \item We present RS-MR, a discovered reduced-state matrix optimizer for transformer pretraining, and demonstrate that it achieves better native multi-stage validation performance than strong baselines while preserving a favorable memory--performance tradeoff.
\end{itemize}

\section{Related Work}

\subsection{Optimizers for Transformer Pretraining}

Optimizer design has been central to large-scale neural network training. Classical first-order methods such as SGD with momentum and Nesterov acceleration provide simple and robust update rules, while adaptive methods such as Adam and AdamW are widely used for transformer pretraining due to their stability and ease of tuning \citep{kingma2014adam, loshchilov2019decoupled}. Memory-efficient variants such as Adafactor reduce second-moment memory costs for large language models \citep{shazeer2018adafactor}, while recent optimizers such as Lion, Sophia, Shampoo-style preconditioners, and Muon-style matrix optimizers explore sign-based updates, curvature approximations, and matrix-structured normalization \citep{zhao2024galore, liu2023sophia, vyas2024soap, zhang2024adam, liu2025muon}.

Despite these advances, most optimizers are still manually designed and tuned through extensive empirical validation, requiring expert intuition about update geometry, state design, numerical stability, and hardware efficiency \citep{shazeer2018adafactor, yuan2024mars,pagliardini2024ademamix}. This challenge is amplified in transformer pretraining by heterogeneous parameter types, long-horizon training dynamics, and the mismatch between short-horizon proxy performance and final training behavior \citep{liu2025cosmos,wen2025fantastic}. \textsc{OPTScientist} differs by treating optimizer design as a searchable typed-program space and validating candidates through native training; the discovered RS-MR optimizer shows that reduced-state matrix mechanisms, such as row-normalized polar directions and soft blockwise gating, can improve performance without heavy second-order states.

\subsection{Automated Optimizer Discovery and Program Search}
\label{subsec:rw_auto_optimizer_discovery}

Another line of work has studied automated discovery of learning rules and optimization algorithms. Learned optimizers parameterize update rules with neural networks and train them through meta-learning \citep{andrychowicz2016learning,romera2024mathematical,novikov2025alphaevolve}, but they are often difficult to interpret, expensive to train, and sensitive to the meta-training distribution. Another direction searches over symbolic or programmatic spaces, where genetic programming, evolutionary algorithms, and AutoML-style systems discover update formulas, neural components, or complete algorithms \citep{koza1994genetic, real2020automl, bello2017neural}; symbolic optimizer discovery has also produced practical methods such as Lion \citep{chen2023symbolic}.

Existing automated discovery methods face a tradeoff between expressivity and reliability. Unconstrained code search is flexible but prone to invalid programs, poor auditability, and brittle implementation behavior \citep{moudgil2025celo,marfinetz2025evolving}, whereas narrowly parameterized spaces are easier to validate but restrict discovery to local variants of known mechanisms \citep{yang2023large,liu2024evolution,ye2024reevo}. \textsc{OPTScientist} addresses this tradeoff by searching inside a typed optimizer DSL that enforces explicit state types, tensor operations, and update semantics while still supporting mechanisms such as momentum, row/column statistics, polar updates, block masking, and structured preconditioning.

\subsection{LLM Agents for Scientific Discovery}

Large language models are increasingly used as agents that reason, call tools, write code, and iteratively improve solutions. Early agentic methods combine language-model reasoning with external actions or feedback, including tool-use agents, ReAct-style reasoning, Reflexion-style self-improvement, and autonomous exploration systems \citep{schick2023toolformer, yao2022react, shinn2023reflexion, wang2023voyager}. Multi-agent frameworks decompose complex tasks into specialized roles for planning, implementation, critique, or verification \citep{wu2024autogen, li2023camel}, and LLM-based scientific discovery systems have been used to generate hypotheses, propose code, search over programs, and improve mathematical or algorithmic constructions \citep{romera2024mathematical, lu2024ai,schmidgall2025agent}.

\textsc{OPTScientist} is related to this line of work but focuses on the constrained and empirically grounded setting of optimizer design for transformer pretraining. Rather than allowing agents to freely edit arbitrary code, it assigns each role a specific scientific function: the \textsc{Theorist} proposes hypotheses, the \textsc{Designer} instantiates DSL programs, the \textsc{Engineer} compiles and evaluates them, and the \textsc{Reviewer} scores candidates and decides whether the DSL should be minimally extended. Compared with general-purpose LLM agents, \textsc{OPTScientist} emphasizes compiler-backed validity, native training evaluation, explicit search memory, and controlled DSL evolution.

\section{Preliminaries}

\paragraph{Optimizers as modular update programs.}
Modern optimizers differ substantially in their surface forms, but many of them can be understood as different ways of constructing the update tensor in $\theta_{t+1}=\theta_t-\eta_t U_t$. For example, SGD chooses the raw gradient or momentum as the update direction, Adam-style methods add adaptive scaling through second-moment statistics, Lion uses sign-based directions, Muon applies matrix-orthogonalized directions, and Shampoo-style methods introduce structured preconditioning. We use this observation to define a structured optimizer design space: an optimizer update can be decomposed into reusable modules for direction, scaling, preconditioning or geometry, regularization, state, and parameter grouping. This decomposition is not intended to be a unique mathematical factorization; rather, it provides a practical coordinate system for searching over optimizer mechanisms.

\paragraph{Optimizer DSL.}
\textsc{OPTScientist} represents each candidate optimizer as a program in a typed domain-specific language (DSL). Instead of asking a language model to generate arbitrary Python optimizer code, the system searches over compact DSL programs composed of metadata declarations, hyperparameter declarations, state updates, intermediate expressions, and a final update expression. At a high level, a DSL program specifies how to compute $U_t$ from the current gradient, parameter tensor, hyperparameters, and persistent optimizer states. This design makes optimizer discovery closer to program synthesis over a structured scientific language than to unconstrained code generation.

\paragraph{Atomic optimizer modules.}
The DSL exposes optimizer building blocks at several levels of abstraction. Direction modules include raw gradients, momentum, sign directions, normalized directions, and matrix-polar directions. Scaling modules include constant scaling, RMS-like scaling, adaptive second-moment scaling, global damping, and norm-based rescaling. Geometry and preconditioning modules include diagonal transformations, row or column normalization, blockwise operations, low-rank or sketched transformations, and matrix-factor or polar operations. Regularization and stabilization modules include decoupled weight decay, clipping, update sanitization, and structured masks. The DSL also supports different state granularities, ranging from scalar and parameter-shaped states to row-wise, column-wise, blockwise, and matrix-factor states.

\paragraph{Search space.}
The resulting search space is combinatorial: a candidate optimizer is formed by selecting state variables, choosing update modules, composing tensor operations, setting hyperparameters, and deciding the granularity at which each mechanism is applied. This space contains many existing optimizer families as special cases, including momentum SGD, RMSProp, AdamW, Lion-like sign optimizers, Muon-like matrix optimizers, and Shampoo-style structured preconditioners. More importantly, it also contains hybrid mechanisms that combine motifs from different families, such as matrix-normalized directions with lightweight blockwise gating. This makes the DSL expressive enough for discovery while still more constrained and auditable than arbitrary source-code search.

\paragraph{Compiler-backed validity.}
A DSL candidate is parsed and checked before it is evaluated in training. The compiler enforces coarse tensor typing, operator constraints, and compatibility between the final update and the target parameter group. Invalid programs are rejected before expensive training runs. In our transformer pretraining experiments, discovered DSL optimizers are mainly applied to two-dimensional matrix parameters, while non-matrix parameter groups are handled by the baseline optimizer in the hybrid training stack. This keeps the discovery target focused on matrix-structured optimizer design, where operations such as row normalization, blockwise gating, and polar directions are most natural.

\section{Method}

\begin{figure*}[t]
\centering
\includegraphics[width=0.98\textwidth]{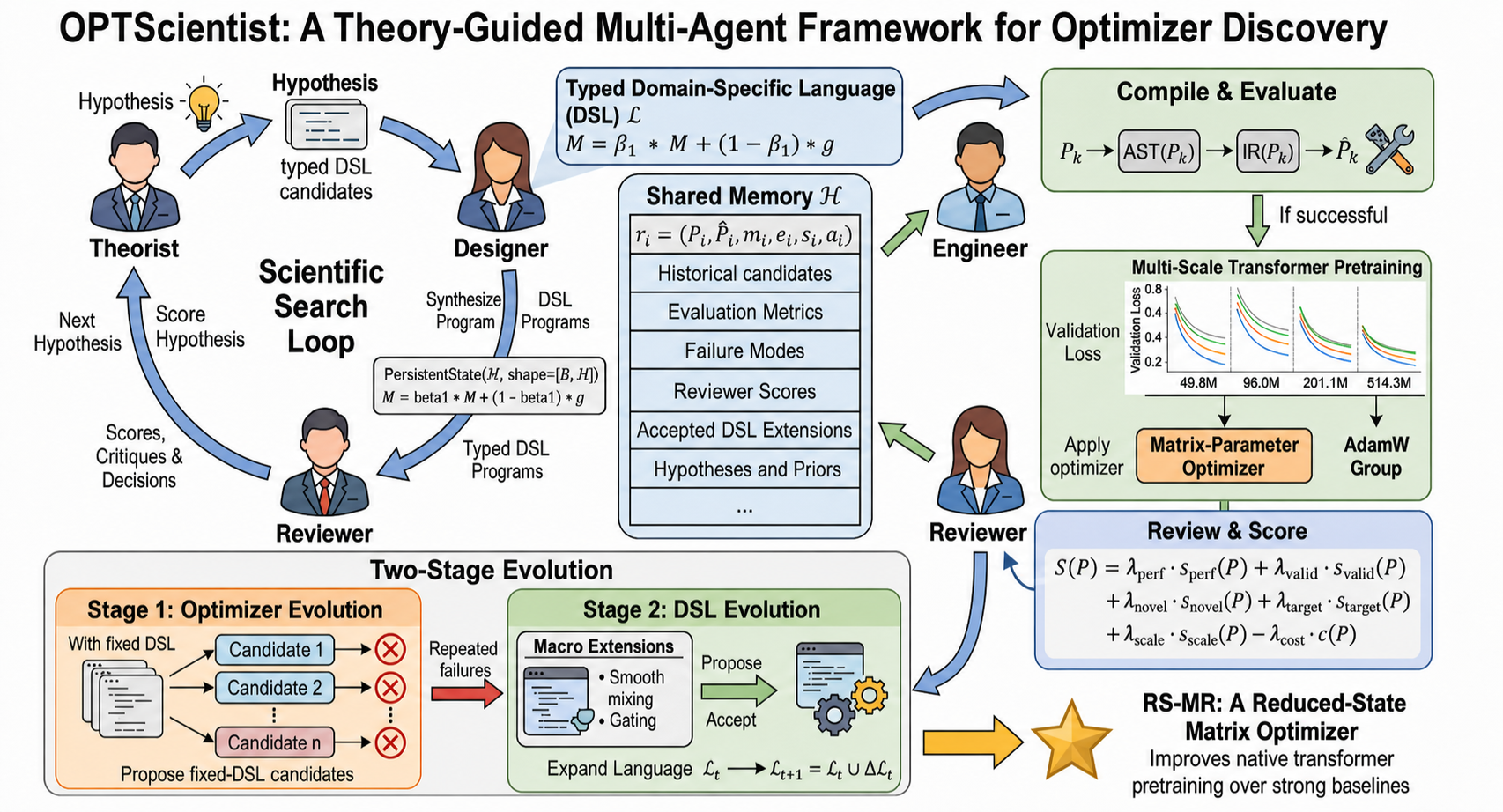}
\caption{
Overview of \textsc{OPTScientist}. The framework discovers optimizer programs through a theory-guided multi-agent loop. A \textsc{Theorist} proposes optimizer hypotheses, a \textsc{Designer} instantiates them as typed DSL programs, an \textsc{Engineer} compiles and evaluates candidates through transformer pretraining, and a \textsc{Reviewer} scores candidates and updates shared memory. The system alternates between fixed-DSL optimizer evolution and conservative DSL evolution through safe macro extensions.}
\label{fig:optscientist_framework}
\end{figure*}

We introduce \textsc{OPTScientist}, a theory-guided multi-agent framework for discovering optimizer programs for transformer pretraining. Instead of asking a language model to write arbitrary Python optimizer code, \textsc{OPTScientist} searches over a typed optimizer DSL. Each candidate is a compact program specifying optimizer states, tensor operations, hyperparameters, and a final update expression. The program is compiled into an executable PyTorch optimizer, evaluated in a native training pipeline, reviewed by a role-specialized agent, and stored in a shared search memory for later rounds.

\subsection{Multi-Agent Search over Typed Optimizer Programs}

Given a typed DSL $\mathcal{L}$, the goal is to find an optimizer program $P\in\mathcal{L}$ that performs well under a target pretraining protocol. The system maintains a shared memory $\mathcal{H}$ containing prior candidates, evaluation outcomes, failure modes, reviewer critiques, and accepted DSL extensions. Each search round follows the loop
\emph{hypothesis} $\rightarrow$ \emph{DSL candidates} $\rightarrow$ \emph{compiled optimizers} $\rightarrow$ \emph{training evaluation} $\rightarrow$ \emph{review} $\rightarrow$ \emph{memory update}.

\begin{wrapfigure}{r}{0.50\textwidth}
\vspace{-1.2em}
\centering
\begin{minipage}{0.48\textwidth}
\captionsetup{type=algorithm}
\caption{\textsc{OPTScientist} search loop.}
\label{alg:optscientist}
\footnotesize
\begin{algorithmic}[1]
\Require DSL $\mathcal{L}_0$, compiler $\mathcal{C}$, evaluator $\mathcal{E}$, memory $\mathcal{H}_0$
\State $\mathcal{L}\!\leftarrow\!\mathcal{L}_0$, $\mathcal{H}\!\leftarrow\!\mathcal{H}_0$
\While{budget remains}
    \State $z \leftarrow \textsc{Theorist}(\mathcal{H},\mathcal{L})$
    \State $\mathcal{P} \leftarrow \textsc{Designer}(z,\mathcal{H},\mathcal{L})$
    \For{$P \in \mathcal{P}$}
        \State $(b,\widehat{P},\epsilon)\leftarrow\textsc{CompileOrRepair}(P,\mathcal{C})$
        \State $m \leftarrow \mathcal{E}(\widehat{P})$ if $b$ else $\textsc{Failure}(\epsilon)$
        \State store $(P,\widehat{P},m,\epsilon)$
    \EndFor
    \State $(S,a)\leftarrow\textsc{Reviewer}(\mathcal{P},\mathcal{H})$
    \State $\mathcal{H}\leftarrow\textsc{UpdateMemory}(\mathcal{H},\mathcal{P},S)$
    \If{$a=\textsc{TriggerDSLEvolution}$}
        \State $\Delta\mathcal{L}\leftarrow\textsc{ProposeMacro}(\mathcal{H},\mathcal{L})$
        \State $\mathcal{L}\leftarrow\mathcal{L}\cup\Delta\mathcal{L}$ if accepted
    \EndIf
\EndWhile
\State \Return best final-validation candidate in $\mathcal{H}$
\end{algorithmic}
\end{minipage}
\vspace{-1.0em}
\end{wrapfigure}

The four agents implement different scientific roles. The \textsc{Theorist} turns prior knowledge and recent failures into optimizer hypotheses, such as using row normalization, blockwise gating, or matrix-geometric update directions. The \textsc{Designer} converts these hypotheses into multiple DSL programs and enforces diversity across state choices, normalization paths, and update compositions. The \textsc{Engineer} compiles, optionally repairs, and evaluates candidates using the fixed compiler and evaluator; it does not invent execution semantics. The \textsc{Reviewer} scores candidates by validity, novelty, empirical performance, target-stage progress, scalability, and resource cost, and decides whether the search should continue within the current DSL or trigger DSL evolution.

This role decomposition differs from ordinary evolutionary search. Candidate variation is not an unguided mutation over source code, but a hypothesis-conditioned synthesis step over typed optimizer programs. Selection is also not based only on short-horizon validation loss: the reviewer incorporates compiler validity, target-stage reachability, cross-scale behavior, memory overhead, and repeated failure patterns.

\subsection{Two-Stage Evolution}

\textsc{OPTScientist} alternates between two stages. In \textbf{Stage~1}, the DSL is fixed and the system searches over optimizer programs $P\in\mathcal{L}$. Each round proposes several candidates, compiles and evaluates them, reviews the results, and updates the shared memory. This stage is the main optimizer-discovery loop.

In \textbf{Stage~2}, the system performs conservative DSL evolution. When repeated failures or plateaus suggest that the current language cannot express a promising mechanism, the reviewer may trigger a small extension $\Delta\mathcal{L}$. In the current implementation, these extensions are safe macro-like operators expanded before compilation, such as smooth mixing, gating, or stable interpolation between update branches. The system does not rewrite the compiler or introduce arbitrary grammar changes. This design allows the search space to grow while preserving compiler-backed validation and reproducibility.

\subsection{Compiler-Backed Candidate Realization}

Every candidate proposed by the \textsc{Designer} is represented as DSL text, not executable Python code. The compiler parses the program, checks tensor kinds and operator constraints, lowers it to an intermediate representation, and generates a PyTorch optimizer. Invalid candidates are rejected before expensive training. For example, matrix-only operators must be applied to matrix-like tensors, row or block states must be compatible with two-dimensional parameters, and the final update must match the target parameter shape.

Compiler-backed realization is central to the framework. It prevents unverifiable code generation, makes candidates comparable as concise typed programs, and ensures that evaluation depends on executable optimizer behavior rather than language-model self-assessment.

\subsection{Evaluation and Selection}

Candidate optimizers are evaluated through a deterministic multi-scale transformer pretraining pipeline. During search, candidates are promoted across progressively larger stages, with the target objective emphasizing reachability and performance at the largest stage rather than only small-scale proxy score. The evaluator returns validation performance, completed stages, failed stages, cross-scale score, optimizer-state memory, and other runtime artifacts.

The reviewer converts these raw metrics into an objective-aware selection score. In simplified form, $S(P)$ rewards empirical performance, validity, novelty, target-stage progress, and cross-scale robustness, while penalizing memory overhead, runtime overhead, numerical instability, duplicate behavior, and evaluator failure. This score guides parent selection and memory updates.

Our training setup uses hybrid parameter grouping. The discovered DSL optimizer is applied to two-dimensional transformer matrix parameters, where structured matrix updates are most natural. Embeddings, language-model heads, vector parameters, and scalar parameters are optimized by AdamW. Thus, the discovered object should be interpreted as a matrix-parameter optimizer embedded in a hybrid pretraining stack, not necessarily a full-model replacement for AdamW.

\subsection{Final Validation}
Search-time proxy evaluations are useful for exploration but can favor brittle or memory-heavy candidates. Therefore, \textsc{OPTScientist} distinguishes proxy-best candidates from final-validation winners. The optimizer promoted as the main artifact is selected by native long-horizon validation rather than by proxy score alone. This distinction is important in our experiments: the final discovered RS-MR optimizer is not merely the short-horizon proxy winner, but the candidate family that provides the best long-horizon validation and resource tradeoff.

\section{Discovered Optimizer: RS-MR}

\textsc{OPTScientist} searches over many typed optimizer programs, but the strongest final design found by the full \textsc{OPTScientist} discovery pipeline is a reduced-state matrix optimizer. We call it \textbf{Reduced-State MAGMA-RowNorm} (\textbf{RS-MR}). RS-MR is not chosen as the best short-horizon proxy candidate. It is selected after the full pipeline of hypothesis generation, DSL instantiation, compiler-backed evaluation, and reviewer-guided final validation. This distinction is important: the optimizer promoted in this paper is the one that performs best under native long-horizon transformer pretraining, not merely under proxy search.

\begin{wrapfigure}{r}{0.50\textwidth}
    \vspace{-1.0em}
    \centering
    \begin{minipage}{0.48\textwidth}
    \captionsetup{type=algorithm}
    \caption{\textsc{RS-MR} update rule.}
    \label{alg:rsmr}
    \footnotesize
    \begin{algorithmic}[1]
    \Require gradient $g_t$, momentum $m_{t-1}$, block gate $s_{t-1}$
    \State $m_t \leftarrow \beta_1 m_{t-1} + (1-\beta_1)g_t$
    \State $s_t \leftarrow 0.9s_{t-1} + 0.1\sigma(\operatorname{block\_cossim}(m_t,g_t)/\tau)$
    \State $u_t \leftarrow \operatorname{polar}(\operatorname{row\_norm}(m_t))$
    \State $\alpha_t \leftarrow 1 + 0.3\sigma(\operatorname{rms}(m_t^2)/0.1)$
    \State $\widetilde{u}_t \leftarrow u_t / \operatorname{clip}_{\gamma}(\alpha_t)$
    \State $a_t \leftarrow \sigma(\operatorname{block\_cossim}(\widetilde{u}_t,g_t)/0.15)$
    \State \Return $U_t \leftarrow (s_t \odot a_t)\odot \widetilde{u}_t$
    \end{algorithmic}
    \end{minipage}
    \vspace{-1.0em}
\end{wrapfigure}

RS-MR maintains only two optimizer states: a momentum matrix $m_t$ and a compact blockwise gating state $s_t$. Its update rule is summarized in Algorithm~\ref{alg:rsmr}. The design combines three main mechanisms. First, it uses a Muon-like matrix direction, but applies row-wise normalization before the polar operation. This ``normalize-then-orthogonalize'' step reduces row-scale imbalance before forming the matrix-geometric update. Second, it applies a lightweight RMS-based damping factor to prevent excessive update magnitudes without introducing a full second-moment tensor. Third, it uses a MAGMA-style soft block gate: the slow gate $s_t$ tracks historical blockwise alignment between momentum and gradient, while the fast gate $a_t$ measures instantaneous alignment between the candidate update and the current gradient.

The resulting update is selective rather than uniform. RS-MR preferentially updates blocks that are both historically reliable and currently aligned, while suppressing unstable blocks. This improves over plain Muon not by adding a heavy second-order preconditioner, but by adding low-cost structured update selection. In our training protocol, RS-MR is applied only to two-dimensional transformer matrix parameters; embeddings, value embeddings, language-model heads, vector parameters, and scalar parameters are optimized by AdamW. Therefore, RS-MR should be interpreted as a discovered matrix-parameter optimizer inside a hybrid transformer pretraining stack.

\section{Experiments}

We evaluate whether the optimizer discovered by \textsc{OPTScientist} transfers from search-time proxy evaluation to native long-horizon transformer pretraining. The main question is whether RS-MR, selected by the full \textsc{OPTScientist} discovery pipeline in Section Method, remains effective in the real \textsc{NanoChat} training environment and improves the performance--memory tradeoff over strong baselines~\citep{nanochat}.

\subsection{Dataset and Benchmark}

All optimizer variants are trained on the FineWeb-Edu 100B pretraining corpus, using the Karpathy reshuffled version hosted as \texttt{fineweb-edu-100b-shuffle}. The corpus is tokenized on the fly with the \textsc{NanoChat} tokenizer with vocabulary size $32768$ and packed using BOS-aligned best-fit cropping. In our local environment, 370 shards are available: shards 0--368 are used for training and shard 369 is used for validation. This is more than sufficient for the 3000-step benchmark, and all optimizer variants are trained on the same data prefix, making the comparison controlled.

Our main validation benchmark is the native d21 \textsc{NanoChat} setting. The model has 21 layers, model dimension $1408$, 11 attention heads with head dimension $128$, context length $2048$, and vocabulary size $32768$. It contains $1{,}099{,}370{,}314$ total parameters, of which $545{,}722{,}144$ are non-embedding scaling parameters. Each run uses 3000 optimization steps with total batch size $1{,}048{,}576$ tokens, corresponding to approximately $3.15$B training tokens. Validation is performed every 250 steps on a held-out validation slice of approximately 20M tokens. The metric is validation bits-per-byte (BPB), where lower is better.

\subsection{Baselines and Evaluation Protocol}

We compare RS-MR against representative first-order, adaptive, and matrix-oriented optimizers, including AdamW, Muon, Sophia, RMSProp, Lion, and SGD with momentum, as well as other top candidates discovered during search. All 16 optimizer variants in the native d21 comparison share the same model, data pipeline, tokenizer, batch size, number of steps, and evaluation schedule. Thus, performance differences primarily reflect optimizer behavior rather than changes in data or training configuration.

RS-MR is first discovered through the \textsc{OPTScientist} search process and then re-evaluated in the native d21 benchmark. This final validation is separate from proxy search. The goal is to avoid selecting candidates that overfit short-horizon proxy settings or rely on expensive optimizer states that do not provide a favorable long-horizon tradeoff.

\subsection{Main Results}

Table~\ref{tab:d21_main_results} reports the main native d21 results. RS-MR achieves the best final validation BPB, reaching $0.798375$. It outperforms Muon, which reaches $0.802949$, by $0.004574$ BPB, a relative improvement of about $0.57\%$. It also substantially outperforms AdamW, which reaches $0.898415$ BPB.

\begin{table}[t]
    \centering
    \caption{Native d21 \textsc{NanoChat} results
    (3000 steps, $1.099$\,B parameters, $\sim$3.15\,B training tokens
    on FineWeb-Edu 100B; lower validation BPB is better).
    RS-MR improves over Muon while adding only a small
    optimizer-state memory overhead, and clearly beats all
    single-state and standard two-state baselines.}
    \label{tab:d21_main_results}
    \small
    \begin{tabular}{lccc}
    \toprule
    Optimizer & Final Val BPB $\downarrow$ & Gap to RS-MR & State Memory \\
    \midrule
    RS-MR (ours)   & $\mathbf{0.798375}$ & ---          & $1913.21$\,MB \\
    Muon           & $0.802949$          & $+0.004574$  & $1905.76$\,MB \\
    Sophia         & $0.877353$          & $+0.078978$  & $3811.53$\,MB \\
    AdamW          & $0.898415$          & $+0.100040$  & $3811.53$\,MB \\
    RMSProp        & $0.901605$          & $+0.103230$  & $1905.76$\,MB \\
    Lion           & $0.920101$          & $+0.121726$  & $1905.76$\,MB \\
    SGD+Momentum   & $1.149398$          & $+0.351023$  & $1905.76$\,MB \\
    \bottomrule
    \end{tabular}
\end{table}

The memory overhead over Muon is only $7.45$ MB, or approximately $0.39\%$. Therefore, the gain does not come from a substantially larger optimizer state. Instead, RS-MR improves the Muon-like update by adding row-normalized matrix directions and compact blockwise update selection.

\subsection{Training Dynamics and Pareto Tradeoff}

\begin{figure*}[t]
\centering
\begin{minipage}{0.49\textwidth}
    \centering
    \includegraphics[width=\linewidth]{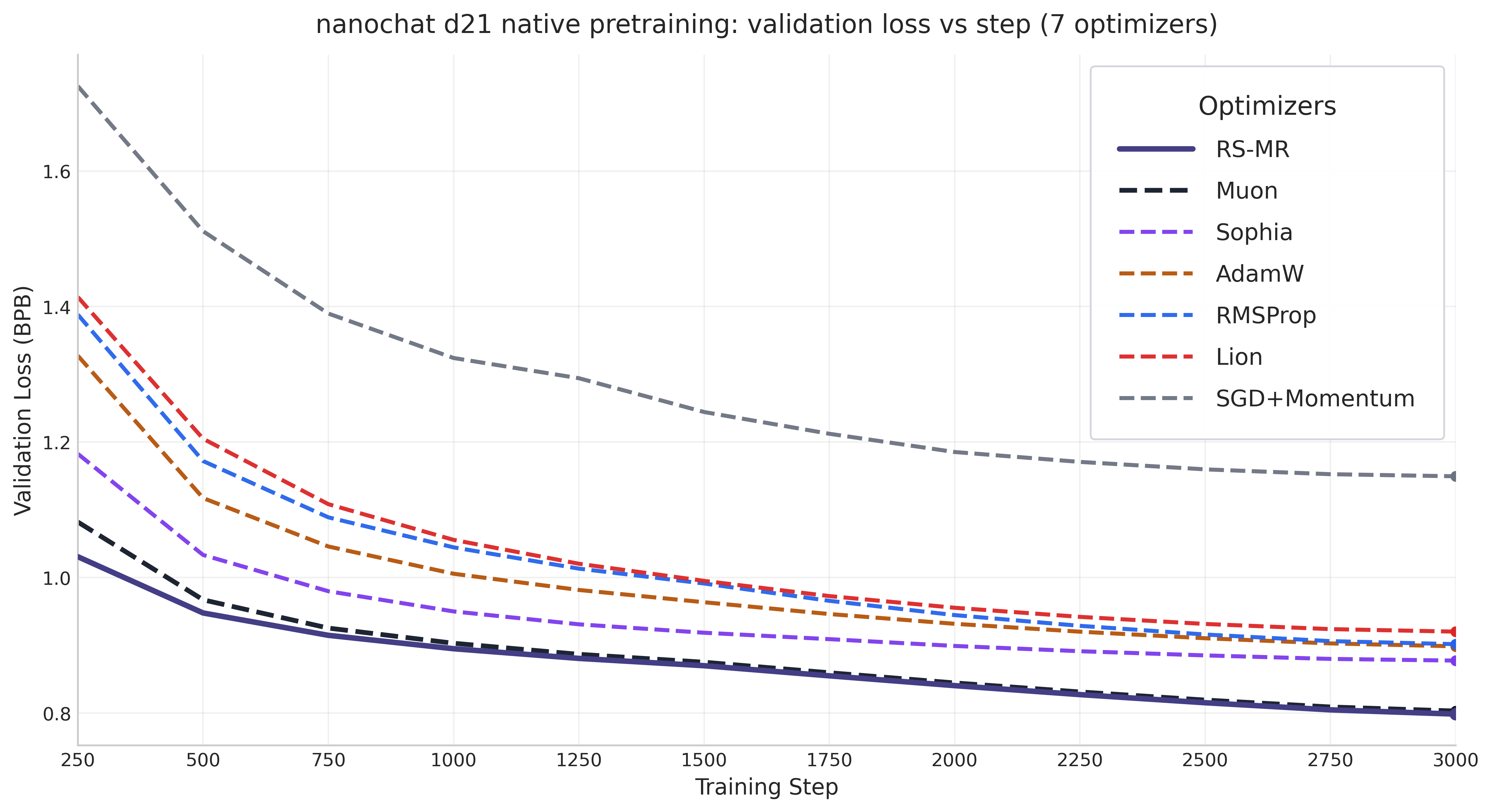}
    \caption*{(a) Validation BPB.}
\end{minipage}
\hfill
\begin{minipage}{0.49\textwidth}
    \centering
    \includegraphics[width=\linewidth]{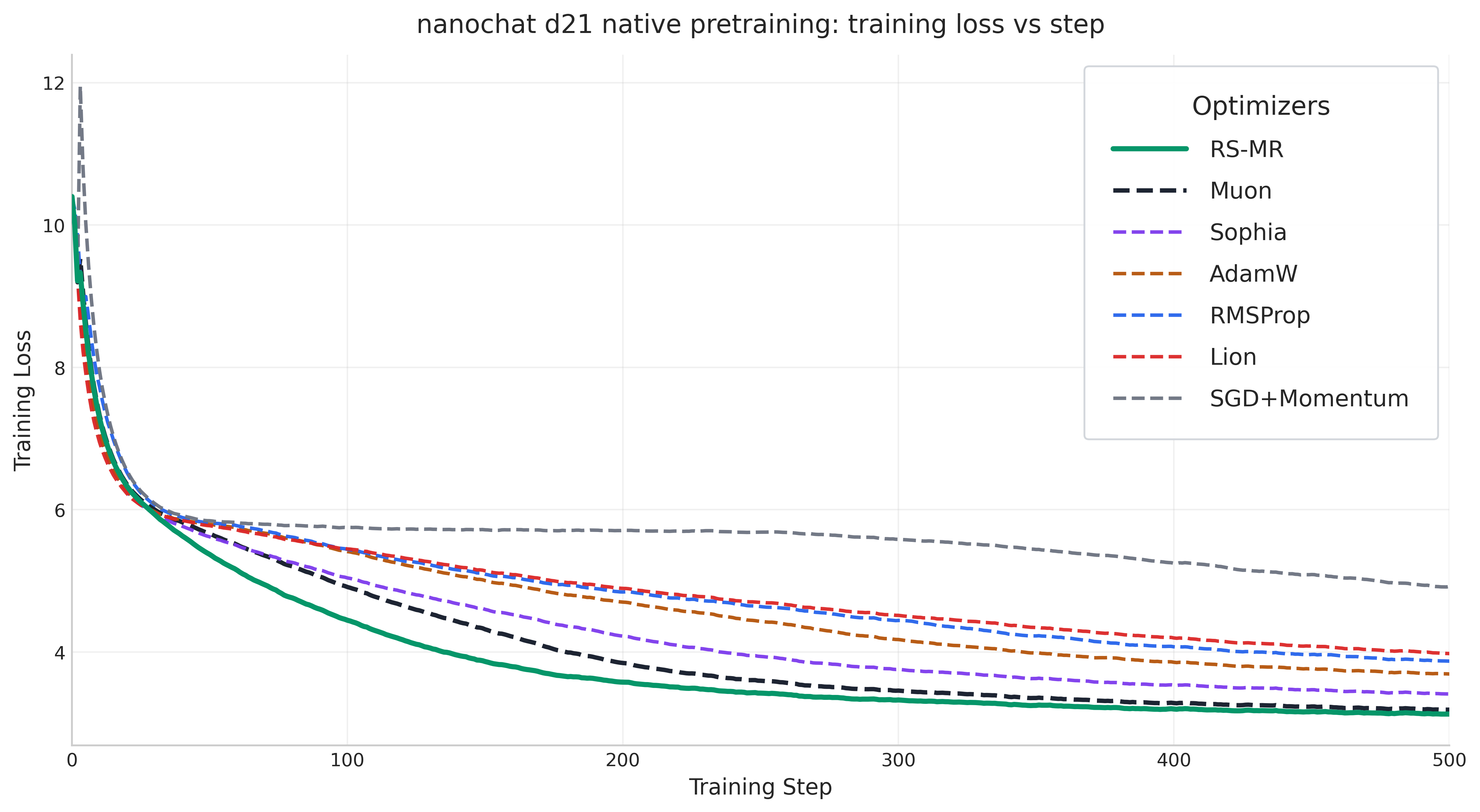}
    \caption*{(b) First-500-step training loss.}
\end{minipage}
\caption{Native d21 \textsc{NanoChat} training dynamics. RS-MR maintains a consistent advantage over Muon throughout the 3000-step validation trajectory and improves early training loss.}
\label{fig:d21_curves}
\end{figure*}

Figure~\ref{fig:d21_curves} shows the validation and training curves. RS-MR is better than Muon at every validation checkpoint from 250 to 3000 steps, indicating that the improvement is not a final-step artifact. The early training curve also shows that RS-MR improves the initial optimization trajectory in the native \textsc{NanoChat} environment.

\begin{figure*}[t]
\centering
\includegraphics[width=0.95\textwidth]{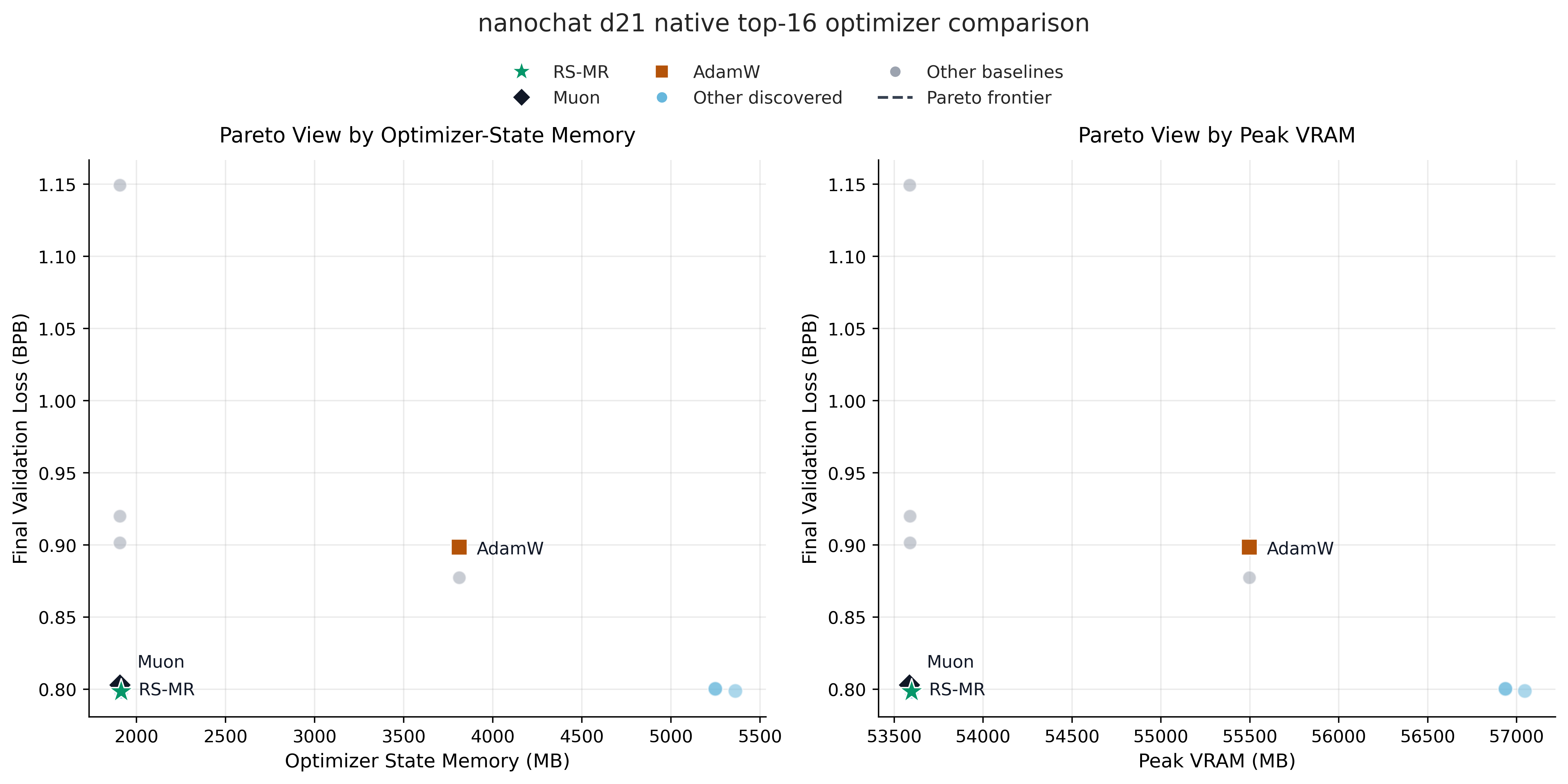}
\caption{Top-16 optimizer comparison on the native d21 benchmark. RS-MR achieves the best validation BPB while staying close to Muon in optimizer-state memory and peak VRAM.}
\label{fig:d21_pareto}
\end{figure*}

Figure~\ref{fig:d21_pareto} compares the top-16 optimizer variants in terms of final validation BPB, optimizer-state memory, and peak VRAM. RS-MR lies in the favorable Pareto region: it achieves the best final validation BPB while remaining close to Muon in both optimizer-state memory and peak VRAM. Some discovered candidates obtain competitive BPB, but they require substantially larger states or higher VRAM. This confirms that RS-MR is not only the best final-validation optimizer, but also the most attractive performance--resource tradeoff among the evaluated candidates.

\subsection{Summary}
\label{subsec:exp_summary}

The native d21 benchmark shows that the optimizer selected by \textsc{OPTScientist}'s full discovery pipeline transfers to the real \textsc{NanoChat} pretraining setting. RS-MR achieves the best final validation BPB, improves over Muon throughout training, substantially outperforms AdamW, and adds only $0.39\%$ optimizer-state memory over Muon. These results support the central claim that typed, compiler-validated optimizer discovery can find practical optimizer mechanisms that improve long-horizon transformer pretraining without relying on heavy optimizer states.

\section{Conclusion and Limitations}
\label{sec:con_lim}
We presented \textsc{OPTScientist}, a theory-guided multi-agent framework for discovering typed optimizer programs for transformer pretraining. By searching over a typed DSL rather than unrestricted optimizer code, \textsc{OPTScientist} combines language-model reasoning, compiler-backed validation, native training evaluation, and reviewer-guided refinement into an auditable optimizer discovery workflow.

The strongest discovered design is RS-MR, a reduced-state matrix optimizer with row-normalized polar directions, lightweight RMS damping, and MAGMA-style soft block gating. On the native 21-layer \textsc{NanoChat} benchmark, RS-MR achieves the best final validation BPB among the evaluated optimizers, consistently improves over Muon, and adds only a small optimizer-state memory overhead. These results show that structured typed-program search can discover practical optimizer mechanisms beyond short-horizon proxy winners.

(Limitations) Our current evidence is mainly from transformer pretraining, especially the native 21-layer \textsc{NanoChat} benchmark. Broader validation on larger models, longer training horizons, more datasets, and other architectures is needed before making general claims about optimizer superiority.

RS-MR is also a matrix-parameter optimizer inside a hybrid training stack: non-matrix parameters are still optimized by AdamW. In addition, the current DSL-evolution stage is conservative and supports only safe macro-like extensions rather than open-ended compiler or language invention. Finally, the search still depends on seed programs, prompt priors, evaluator design, and compute, and the theoretical reason why RS-MR improves over Muon remains an important direction for future work.

\newpage
\bibliographystyle{plainnat}
\bibliography{ref}

\newpage
\appendix
\setcounter{page}{1}
\vbox{
    \hrule height 4pt
    \vskip 0.25in
    \vskip -\parskip%
    \centering
    {\LARGE\bf OPTScientist: Multi-Agent Discovery of Typed
Optimizer Programs for Transformer Pretraining (Appendix)}
    \vskip 0.29in
    \vskip -\parskip
    \hrule height 1pt
}

\appendix

\section{Unified Optimizer Design Space}
\label{app:optimizer_design_space}

This appendix gives a more detailed description of the optimizer design space used to motivate our DSL. The main paper only introduces the high-level idea: optimizers are represented as modular update programs. Here we describe the modules, state structures, granularity choices, and compiler-backed DSL representation in more detail.

\subsection{General Update Form}
\label{app:general_update_form}

We consider neural network training with parameters $\theta_t$, stochastic gradient $g_t=\nabla_\theta \ell(\theta_t;\xi_t)$, learning rate $\eta_t$, and optimizer update tensor $U_t$. The generic update is
\begin{equation}
    \theta_{t+1}
    =
    \theta_t
    -
    \eta_t U_t .
\end{equation}
The central design question is how to construct $U_t$ from the current gradient, parameter tensor, and optimizer history.

We view many optimizers as composing four conceptual modules:
\begin{equation}
    U_t
    =
    D_t(g_t,s_t)
    \odot
    S_t(g_t,s_t)
    \odot
    P_t(g_t,s_t)
    +
    R_t(\theta_t),
    \label{eq:appendix_modular_update}
\end{equation}
where $s_t$ denotes optimizer states. The direction module $D_t$ determines the descent signal, the scaling module $S_t$ controls update magnitude, the preconditioning or geometry module $P_t$ changes the update geometry, and the regularization module $R_t$ captures weight decay, clipping, masks, or constraints. This decomposition is not unique, but it is useful for organizing the search space.

\subsection{Direction Space}
\label{app:direction_space}

The direction module determines where the optimizer moves. Representative choices include:
\begin{align}
    D_t &= g_t
    && \text{raw gradient}, \\
    m_t &= \beta m_{t-1} + (1-\beta)g_t,\quad D_t=m_t
    && \text{momentum}, \\
    D_t &= \operatorname{sign}(m_t)
    && \text{sign direction}, \\
    D_t &= \frac{g_t}{\|g_t\|+\epsilon}
    && \text{normalized gradient}, \\
    D_t &= g_t-g_{t-1}
    && \text{gradient difference}, \\
    D_t &= H_t^{-1}g_t
    && \text{second-order direction}.
\end{align}
Matrix optimizers introduce additional direction choices. For a two-dimensional matrix parameter, a momentum matrix $M_t$ may be transformed by a polar or orthogonalization operator:
\begin{equation}
    D_t = \operatorname{polar}(M_t).
\end{equation}
This captures the core motif behind Muon-like update rules.

\subsection{Scaling Space}
\label{app:scaling_space}

The scaling module determines how far to move along the chosen direction. Common choices include constant scaling, RMS scaling, second-moment scaling, norm-based scaling, signal-to-noise-ratio-like scaling, and curvature-based damping. For example, Adam-style methods maintain
\begin{equation}
    v_t
    =
    \beta_2 v_{t-1}
    +
    (1-\beta_2)g_t^2
\end{equation}
and use a scaling factor of the form
\begin{equation}
    S_t
    =
    \frac{1}{\sqrt{v_t}+\epsilon}.
\end{equation}
Other optimizers may use global RMS statistics, layer-wise norms, or scalar damping factors to control the step size without maintaining a full parameter-shaped second-moment state.

\subsection{Preconditioning and Geometry Space}
\label{app:preconditioning_space}

The preconditioning module changes the geometry of the update. Diagonal preconditioners use elementwise statistics and are the basis of Adagrad, RMSProp, and Adam-style methods:
\begin{equation}
    P_t = \operatorname{diag}(p_t).
\end{equation}
Block-diagonal and Kronecker-factored preconditioners approximate curvature with structured matrices:
\begin{equation}
    P_t
    =
    \operatorname{blockdiag}(H_1^{-1},H_2^{-1},\ldots),
    \qquad
    H^{-1} \approx A^{-1}\otimes B^{-1}.
\end{equation}
Low-rank preconditioners approximate curvature with a small subspace:
\begin{equation}
    H^{-1}
    \approx
    U\Lambda^{-1}U^\top .
\end{equation}
Matrix-geometric methods instead transform the update direction through normalization or polar operations. Examples include row normalization, column normalization, blockwise transformations, and approximate polar decomposition.

\subsection{Regularization and Stabilization}
\label{app:regularization_space}

Regularization and stabilization modules include L2 regularization, decoupled weight decay, clipping, trust-region-like constraints, and masks. Classical L2 regularization adds a term $R_t(\theta_t)=\lambda\theta_t$ to the update. Decoupled weight decay instead applies shrinkage directly:
\begin{equation}
    \theta_{t+1}
    =
    (1-\eta_t\lambda)\theta_t
    -
    \eta_t U_t .
\end{equation}
Stabilization mechanisms such as clipping and sanitization can be viewed as constraints on the update tensor, for example enforcing $\|U_t\|\leq c$ or replacing invalid numerical values before applying the update. Structured masks can further suppress unreliable coordinates, rows, columns, or blocks.

\subsection{State Space and Granularity}
\label{app:state_granularity}

Optimizer states store information from previous steps. Common states include first moments, second moments, Hessian or curvature estimates, exponential moving averages, gradient-history buffers, and blockwise reliability statistics. These states can have different shapes and memory costs:
\begin{itemize}
    \item scalar states;
    \item vector or parameter-shaped states;
    \item row-wise or column-wise matrix states;
    \item blockwise matrix states;
    \item square matrix-factor states;
    \item global states shared across parameter groups.
\end{itemize}
The update itself can also be applied at different granularities, including parameter-wise, tensor-wise, layer-wise, block-wise, or globally. This choice strongly affects both memory cost and optimization behavior.

\section{Optimizer DSL Details}
\label{app:dsl_details}

\subsection{Program Structure}
\label{app:dsl_program_structure}

The DSL used by \textsc{OPTScientist} is a compact line-based language for defining optimizer programs. A program contains metadata declarations, hyperparameter declarations, state declarations, optional intermediate assignments, and a final update expression:
\begin{verbatim}
meta kind = Mat2D
hparam beta1 = 0.95
hparam tau = 0.15

state m = beta1 * m + (1 - beta1) * grad
u = polar(row_norm(m))
update = u
\end{verbatim}
The final \texttt{update} expression must produce a tensor compatible with the parameter being optimized. The training loop applies the update; the DSL program does not directly mutate model parameters.

\subsection{Metadata and Hyperparameters}
\label{app:dsl_meta_hparam}

The \texttt{meta} fields provide typing or backend information. For example, \texttt{meta kind = Mat2D} indicates that the program is intended for two-dimensional matrix parameters. Other metadata fields can control execution semantics such as masking granularity, state initialization, or numerical safeguards, depending on the backend implementation.

Hyperparameters are declared explicitly:
\begin{verbatim}
hparam lr = 0.02
hparam beta1 = 0.95
hparam eps = 1e-8
\end{verbatim}
The compiler records these declarations and exposes them as scalar values in the generated optimizer.

\subsection{State Declarations}
\label{app:dsl_state_declarations}

The DSL supports several persistent state declarations. The most common declaration is \texttt{state}, which creates a parameter-shaped state:
\begin{verbatim}
state m = beta1 * m + (1 - beta1) * grad
\end{verbatim}
This can express momentum, second moments, or other full-tensor states. The DSL also supports reduced or structured state types:
\begin{itemize}
    \item \texttt{state\_row}: row-wise matrix state;
    \item \texttt{state\_col}: column-wise matrix state;
    \item \texttt{state\_block}: blockwise matrix state;
    \item \texttt{state\_mat}: left square matrix-factor state;
    \item \texttt{state\_rmat}: right square matrix-factor state;
    \item \texttt{state\_scalar}: per-parameter scalar state;
    \item \texttt{global\_state\_scalar}: optimizer-wide scalar state.
\end{itemize}
These declarations allow the search to trade off expressivity and memory. For example, a blockwise state can encode structural update reliability at much lower cost than a full parameter-shaped second-moment state.

\subsection{Operators}
\label{app:dsl_operators}

The DSL operator set includes arithmetic operations, reductions, elementwise nonlinearities, normalization functions, matrix operations, and blockwise functions. Representative operators include:
\begin{itemize}
    \item elementwise operations such as \texttt{square}, \texttt{sqrt}, \texttt{sign}, \texttt{abs}, \texttt{sigmoid}, and \texttt{clip};
    \item scalar reductions such as \texttt{norm}, \texttt{rms}, \texttt{max\_val}, and \texttt{min\_val};
    \item row and column statistics such as \texttt{row\_mean}, \texttt{row\_rms}, \texttt{col\_mean}, and \texttt{col\_rms};
    \item matrix operations such as \texttt{polar}, \texttt{newton\_schulz}, \texttt{transpose}, \texttt{diag}, and \texttt{outer};
    \item block operations such as \texttt{block\_rms}, \texttt{block\_mask}, and \texttt{block\_cossim};
    \item structured transformations such as \texttt{low\_rank}, \texttt{sketch}, \texttt{project\_psd}, and \texttt{block\_inv\_sqrt}.
\end{itemize}
This operator library is designed to cover a broad range of optimizer motifs while remaining statically checkable.

\subsection{Typing and Validation}
\label{app:dsl_typing_validation}

The compiler distinguishes scalar expressions from tensor expressions and assigns coarse tensor kinds to tensor expressions. The main tensor kinds include \texttt{Vec1D}, \texttt{Mat2D}, \texttt{MatRow}, \texttt{MatCol}, \texttt{MatLeftSquare}, \texttt{MatRightSquare}, and a generic \texttt{Any} kind. Operator constraints are checked before evaluation. For example, \texttt{polar} and \texttt{newton\_schulz} require matrix-like tensors; row and column reductions require matrix-compatible inputs; and the final update must be a tensor whose kind is compatible with the target parameter.

The compiler pipeline is:
\begin{equation}
    P
    \xrightarrow{\mathrm{parse}}
    \mathrm{AST}(P)
    \xrightarrow{\mathrm{typecheck}}
    \mathrm{IR}(P)
    \xrightarrow{\mathrm{codegen}}
    \widehat{P},
\end{equation}
where $P$ is the DSL program and $\widehat{P}$ is the generated PyTorch optimizer. This pipeline prevents invalid candidates from reaching the expensive training evaluator and makes discovered optimizers auditable as compact typed programs.

\subsection{Known Optimizers as DSL Programs}
\label{app:known_optimizers_dsl}

Many known optimizers can be expressed as short DSL programs. A simplified momentum optimizer is:
\begin{verbatim}
hparam beta = 0.9
state m = beta * m + (1 - beta) * grad
update = m
\end{verbatim}
A simplified Adam-like update is:
\begin{verbatim}
hparam beta1 = 0.9
hparam beta2 = 0.999
hparam eps = 1e-8

state m = beta1 * m + (1 - beta1) * grad
state v = beta2 * v + (1 - beta2) * square(grad)

update = m / (sqrt(v) + eps)
\end{verbatim}
A simplified Muon-like matrix update is:
\begin{verbatim}
meta kind = Mat2D
hparam beta1 = 0.95

state m = beta1 * m + (1 - beta1) * grad
update = polar(m)
\end{verbatim}
These examples illustrate why the DSL is useful for search: known optimizers become points in the same program space, while new optimizers can be discovered by recombining modules, states, and tensor operations.

\subsection{Hybrid Parameter Grouping}
\label{app:hybrid_parameter_grouping}

In our transformer pretraining experiments, the discovered DSL optimizer is primarily applied to two-dimensional transformer matrix parameters. Other parameter groups, such as token embeddings, value embeddings, language-model heads, vector parameters, and scalar parameters, are optimized by the baseline optimizer used by the training stack. This hybrid grouping is important for both numerical stability and fair evaluation. The discovered object should therefore be interpreted as a matrix-parameter optimizer embedded inside a hybrid training system, rather than as a full-model replacement for every parameter class.

\section{Additional Method Details}
\label{app:method_details}

\subsection{Shared Memory}
\label{app:shared_memory}

The shared memory $\mathcal{H}$ stores all information needed to make the search iterative rather than a collection of independent trials. Each candidate record can be written as
\begin{equation}
    r_i=(P_i,\widehat{P}_i,m_i,e_i,s_i,a_i),
\end{equation}
where $P_i$ is the DSL program, $\widehat{P}_i$ is the compiled optimizer if compilation succeeds, $m_i$ denotes evaluator metrics, $e_i$ denotes compiler or runtime errors, $s_i$ denotes reviewer scores, and $a_i$ denotes agent annotations such as design notes or critiques.

The memory includes historical best candidates, selection-best candidates, recent failures, target-stage failures, plateau indicators, accepted DSL extensions, previous hypotheses, reviewer critiques, candidate lineage, evaluation metrics, stage-completion flags, and optimizer-state memory statistics. The \textsc{Theorist} uses this memory to identify promising mechanisms and recurring failures; the \textsc{Designer} uses it to avoid redundant candidates; the \textsc{Engineer} uses it for cache-aware evaluation and repair; and the \textsc{Reviewer} uses it to detect plateaus and decide whether DSL evolution should be triggered.

\subsection{Agent Interfaces}
\label{app:agent_interfaces}

\paragraph{Theorist.}
The \textsc{Theorist} reads $\mathcal{H}$, the current DSL $\mathcal{L}_t$, recent candidate outcomes, and optional paper summaries. It outputs
\begin{equation}
    z_t^{\mathrm{th}}=\{h_t,\pi_t,\rho_t,q_t,e_t\},
\end{equation}
where $h_t$ is a mechanism hypothesis, $\pi_t$ is a set of search priors, $\rho_t$ is a parent-selection strategy, $q_t$ is a DSL-pressure estimate, and $e_t$ is an optional extension request.

\paragraph{Designer.}
The \textsc{Designer} maps the hypothesis into a set of candidates $\mathcal{P}_t=\{P_{t,1},\ldots,P_{t,K}\}$, where each $P_{t,k}\in\mathcal{L}_t$. Candidates are emitted as DSL text. The system encourages structural diversity across state definitions, tensor operations, normalization rules, preconditioning choices, and gating mechanisms. If LLM generation fails, the system falls back to deterministic DSL templates.

\paragraph{Engineer.}
The \textsc{Engineer} invokes the compiler on each candidate:
\begin{equation}
    \mathrm{Compile}(P_{t,k})
    \rightarrow
    (b_{t,k}^{\mathrm{comp}},\widehat{P}_{t,k},\epsilon_{t,k}^{\mathrm{comp}}),
\end{equation}
where $b_{t,k}^{\mathrm{comp}}$ indicates compile success, $\widehat{P}_{t,k}$ is the compiled optimizer, and $\epsilon_{t,k}^{\mathrm{comp}}$ is the compiler error. If compilation fails, the Engineer may perform a bounded repair step using the compiler error. Successful candidates are passed to the evaluator, which returns metrics $m_{t,k}$.

\paragraph{Reviewer.}
The \textsc{Reviewer} computes candidate scores
\begin{equation}
    \mathbf{s}(P)=
    (s_{\mathrm{novel}},s_{\mathrm{valid}},s_{\mathrm{perf}},
    s_{\mathrm{target}},s_{\mathrm{scale}}),
\end{equation}
measuring structural novelty, compiler/runtime validity, empirical performance, target-stage progress, and cross-scale robustness. The reviewer also emits a stage-2 signal when repeated failures suggest a language-level bottleneck.

\subsection{Selection Score}
\label{app:selection_score}

The reviewer combines metrics into an objective-aware selection score:
\begin{equation}
\begin{aligned}
    S(P)
    =
    &
    \lambda_{\mathrm{perf}} s_{\mathrm{perf}}(P)
    +
    \lambda_{\mathrm{valid}} s_{\mathrm{valid}}(P)
    +
    \lambda_{\mathrm{novel}} s_{\mathrm{novel}}(P)
    \\
    &
    +
    \lambda_{\mathrm{target}} s_{\mathrm{target}}(P)
    +
    \lambda_{\mathrm{scale}} s_{\mathrm{scale}}(P)
    -
    \lambda_{\mathrm{cost}} c(P).
\end{aligned}
\end{equation}
Here $c(P)$ penalizes memory overhead, runtime overhead, numerical instability, duplicate signatures, evaluator failure, and target-stage collapse. The exact weighting can be implementation-dependent, but the intended behavior is fixed: candidates should be valid, novel, stable across scales, capable of reaching the target stage, and resource-efficient.

\subsection{Detailed Compiler Pipeline}
\label{app:compiler_pipeline}

A DSL program contains metadata declarations, hyperparameter declarations, persistent state updates, intermediate expressions, and a final update expression. The compiler pipeline is
\begin{equation}
    P
    \xrightarrow{\mathrm{parse}}
    \mathrm{AST}(P)
    \xrightarrow{\mathrm{typecheck}}
    \mathrm{IR}(P)
    \xrightarrow{\mathrm{codegen}}
    \widehat{P}.
\end{equation}
The type checker distinguishes scalar expressions from tensor expressions and assigns coarse structural kinds to tensors, such as vector, matrix, row-reduced, column-reduced, blockwise, and matrix-factor kinds. Operator constraints are checked before evaluation. For example, \texttt{polar} requires matrix-like tensors, row and column reductions require matrix-compatible inputs, and the final update must be a tensor compatible with the target parameter group.

This pipeline reduces wasted evaluator budget and improves auditability. Instead of inspecting a long generated Python optimizer, one can inspect the concise DSL program and its compiler-produced optimizer.

\subsection{Parameter Grouping}
\label{app:parameter_grouping}

Transformer parameters are heterogeneous. Matrix-oriented operations, such as polar directions, row/column normalization, structured preconditioning, and blockwise gating, are naturally suited for two-dimensional matrices but are not always appropriate for embeddings, output heads, vector parameters, or scalar parameters. Therefore, our evaluation uses hybrid parameter grouping.

Let $\theta=\theta_{\mathrm{mat}}\cup\theta_{\mathrm{aux}}$, where $\theta_{\mathrm{mat}}$ denotes two-dimensional transformer matrix parameters and $\theta_{\mathrm{aux}}$ denotes auxiliary non-matrix parameters. The DSL optimizer updates the matrix group as
\begin{equation}
    \theta_{\mathrm{mat}}^{t+1}
    =
    \theta_{\mathrm{mat}}^t
    -
    \eta U_P(g_{\mathrm{mat}}^t,s^t),
\end{equation}
where $U_P$ is the update compiled from DSL program $P$. The auxiliary group is updated by AdamW:
\begin{equation}
    \theta_{\mathrm{aux}}^{t+1}
    =
    \mathrm{AdamW}(\theta_{\mathrm{aux}}^t,g_{\mathrm{aux}}^t).
\end{equation}
This avoids overstating the scope of the discovered optimizer. The discovered program is a matrix-parameter optimizer inside a hybrid transformer pretraining stack.

\subsection{Multi-Scale Search-Time Evaluation}
\label{app:multiscale_eval}

During search, candidates are evaluated with a multi-scale transformer pretraining protocol. The stages are
\begin{equation}
    \mathcal{M}=
    \{49.8\mathrm{M},96.0\mathrm{M},201.1\mathrm{M},514.3\mathrm{M}\}.
\end{equation}
Each stage uses a fixed training budget and returns validation bits-per-byte, stage-completion flags, failure information, optimizer-state memory, and other metrics. Early-stage scores provide useful feedback, but the target objective emphasizes stability and performance at the largest stage. This discourages candidates that look strong at small scale but fail during promotion.

\subsection{Proxy Search and Final Validation}
\label{app:proxy_final_validation}

The search loop may use cheaper proxy evaluations for exploration, but the main optimizer artifact is selected by final native validation. This separation is necessary because proxy-best candidates can be brittle, memory-heavy, or specialized to short horizons. Final validation filters for long-horizon robustness, memory efficiency, and consistent training behavior. In our experiments, this protocol selects RS-MR as the representative discovered optimizer.

\begin{table}[t]
\centering
\caption{All 16 optimizer variants on the native d21 \textsc{NanoChat}
3000-step benchmark, ranked by final validation BPB.
RS-MR is the same DSL program as OE-1 (\texttt{rank\_008\_\_oe\_20260325\_222424}).
The top 10 OpenEvolve-discovered DSL optimizers (OE-1\,$\sim$\,OE-10) all
outperform every classical baseline. The number of dense optimizer
states is reported as recorded by the DSL compiler; small block-level
states (e.g.\ a $16{\times}16$ MAGMA mask) are counted but contribute
negligibly to memory.}
\label{tab:d21_full_results}
\small
\begin{tabular}{lcccc}
\toprule
Optimizer & Final Val BPB $\downarrow$ & Gap to RS-MR
          & State Mem.\ (MB) & \#States \\
\midrule
RS-MR (=\,OE-1, ours) & $\mathbf{0.798375}$ & ---         & $1913.21$ & 2 \\
OE-2  & $0.798390$ & $+0.000015$ & $1913.21$ & 2 \\
OE-3  & $0.798424$ & $+0.000049$ & $1913.21$ & 2 \\
OE-4  & $0.798433$ & $+0.000058$ & $1913.21$ & 2 \\
OE-5  & $0.798438$ & $+0.000063$ & $1913.21$ & 2 \\
OE-6  & $0.798952$ & $+0.000577$ & $5359.94$ & 3 \\
OE-7  & $0.799940$ & $+0.001565$ & $1915.24$ & 4 \\
OE-8  & $0.800225$ & $+0.001850$ & $5248.28$ & 3 \\
OE-9  & $0.800269$ & $+0.001894$ & $5248.28$ & 3 \\
OE-10 & $0.801518$ & $+0.003143$ & $1913.21$ & 2 \\
\midrule
Muon           & $0.802949$ & $+0.004574$ & $1905.76$ & 1 \\
Sophia         & $0.877353$ & $+0.078978$ & $3811.53$ & 2 \\
AdamW          & $0.898415$ & $+0.100040$ & $3811.53$ & 2 \\
RMSProp        & $0.901605$ & $+0.103230$ & $1905.76$ & 1 \\
Lion           & $0.920101$ & $+0.121726$ & $1905.76$ & 1 \\
SGD+Momentum   & $1.149398$ & $+0.351023$ & $1905.76$ & 1 \\
\bottomrule
\end{tabular}
\end{table}

\end{document}